\documentclass[journal]{IEEEtran} 

\usepackage{amsmath}

\usepackage[dvipsnames]{xcolor}

\usepackage{float}

\usepackage{graphicx} 
\graphicspath{{images/}}
\DeclareGraphicsExtensions{.pdf,.jpeg,.png}

\usepackage[caption=false]{subfig}

\usepackage{multicol}
\usepackage{multirow}

\usepackage{url}

\usepackage{verbatim}

\usepackage{booktabs}
\usepackage{array}

\begin{document}

\title{Identifying Cross-Depicted Historical Motifs}

\author{
    \IEEEauthorblockN{
        \textbf{Vinaychandran~Pondenkandath}\IEEEauthorrefmark{1}\IEEEauthorrefmark{2}, \and
        \textbf{Michele~Alberti}\IEEEauthorrefmark{1}\IEEEauthorrefmark{2}, \and
        \thanks{\IEEEauthorrefmark{1} Both authors contributed equally to this work.}
        \textbf{Nicole Eichenberger}\IEEEauthorrefmark{3}, \and
        \textbf{Rolf~Ingold}\IEEEauthorrefmark{2}, \and
        \textbf{Marcus~Liwicki}\IEEEauthorrefmark{2}\IEEEauthorrefmark{4}
    }\\
    \vspace{0.2cm}
    \IEEEauthorblockA{
        \IEEEauthorrefmark{2}%
        \textit{Document Image and Voice Analysis Group (DIVA)} \\
        University of Fribourg, Bd. de Perolles 90, 1700 Fribourg, Switzerland\\
        \{firstname\}.\{lastname\}@unifr.ch \\
        \vspace{0.15cm}
        \IEEEauthorrefmark{3}%
        \textit{Staatsbibliothek zu Berlin --– Preu{\ss}ischer Kulturbesitz} \\
        Potsdamer Stra{\ss}e 33, 10785 Berlin, Germany\\
        nicole.eichenberger@sbb.spk-berlin.de\\
        \vspace{0.15cm}
        \IEEEauthorrefmark{4}%
        \textit{MindGarage} \\
        University of Kaiserslautern, Erwin-Schr\"odinger-Stra{\ss}e 1, 67663 Kaiserslautern, Germany
    }
}

\markboth{}%
{Alberti \MakeLowercase{\textit{et al.}}: TODO INSERT TITLE HERE}

\maketitle
\thispagestyle{empty}


\begin{abstract}

Cross-depiction is the problem of identifying the same object even when it is depicted in a variety of manners.
This is a common problem in handwritten historical documents image analysis, for instance when the same letter or motif is depicted in several different ways.
It is a simple task for humans yet conventional heuristic computer vision methods struggle to cope with it.
In this paper we address this problem using state-of-the-art deep learning techniques on a dataset of historical watermarks containing images created with different methods of reproduction, such as hand tracing, rubbing, and radiography.
To study the robustness of deep learning based approaches to the cross-depiction problem, we measure their performance on two different tasks: classification and similarity rankings.
For the former we achieve a classification accuracy of 96\,\% using deep convolutional neural networks.
For the latter we have a false positive rate at 95\% true positive rate of 0.11.
These results outperform state-of-the-art methods by a significant margin.

\end{abstract}

\section{Introduction}
\label{toc:intro}

\begin{figure}[!t]
  \centering
  \subfloat[Query image]%
  {\hspace*{8pt}
  \includegraphics[height=3.3cm]{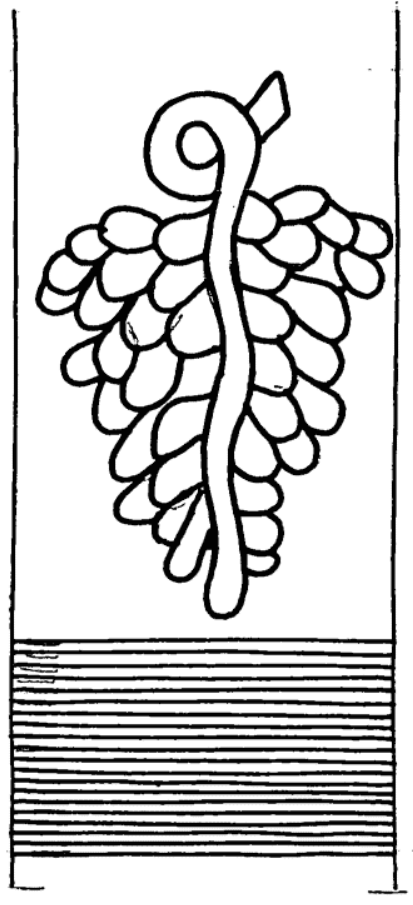}\label{subfig:query}
  \hspace*{8pt}}
  \vfill
  
  \subfloat[Expert result 1]%
  {\includegraphics[height=3.3cm]%
  {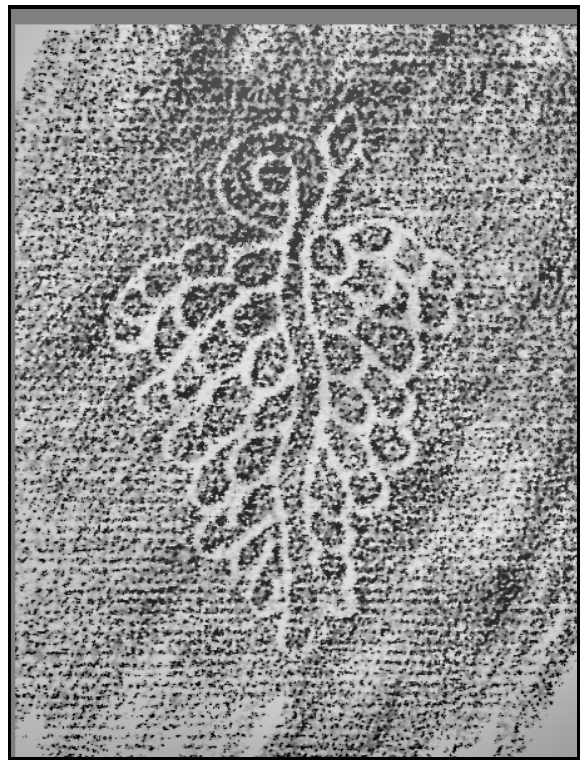}\label{subfig:e1}}
  \hfil
  \subfloat[Expert result 2]%
  {\hspace*{8pt}
  \includegraphics[height=3.3cm]%
  {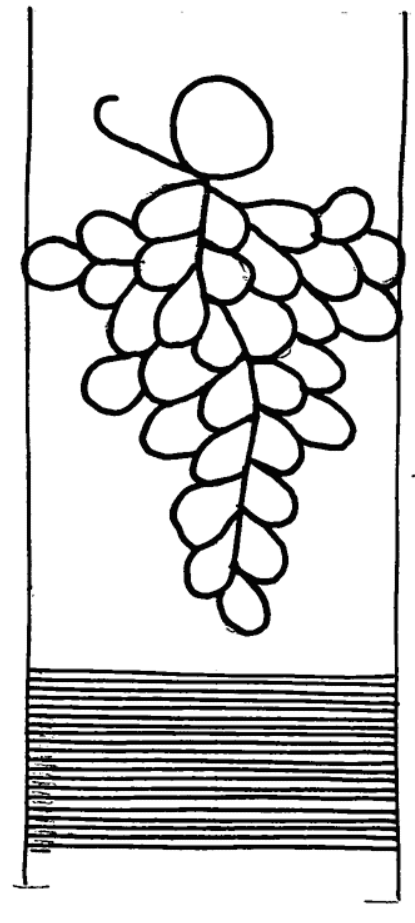}\label{subfig:e2}
  \hspace*{8pt}}
  \hfil
  \subfloat[Expert result 3]%
  {\includegraphics[height=3.3cm]%
  {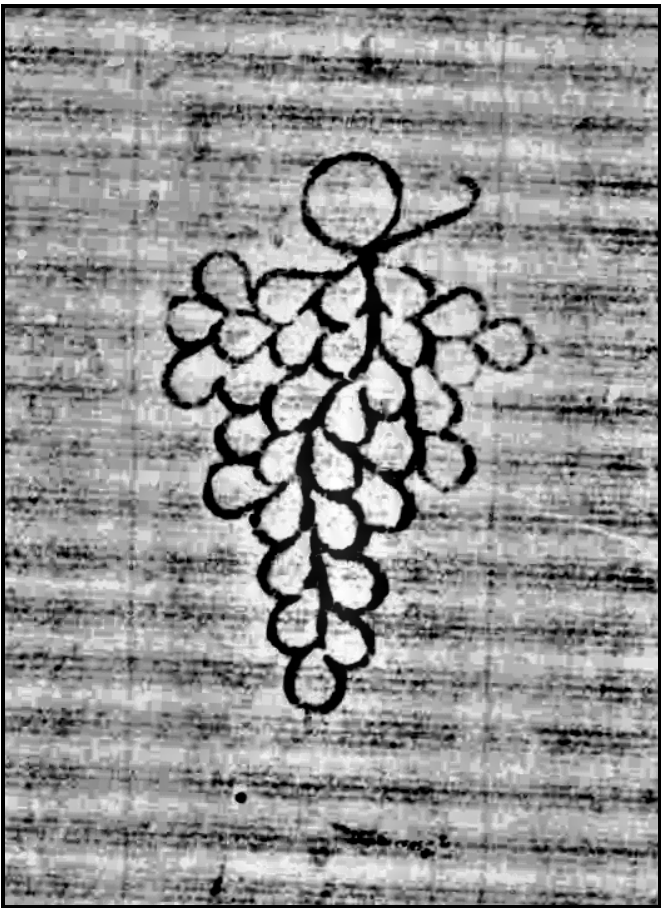}\label{subfig:e3}}
  
  \caption{Example of a query (a) with the expert annotated results (b, c, d). Our system retrieves exactly the same results in the order (c, d, b). Notice that image (a) and (b) only differ by the reproduction method (hand tracing, rubbing), but depict the very same watermark. Unlike previous approaches, our system successfully retrieves it within the top 3 results}
  \label{fig:queried_images}
\end{figure}

Dating historical manuscripts is fundamental for historians and there are numerous attempts in computer science at solving it~\cite{wahlberg2015, he2016codebook, he2016image}.
The identification of the paper's watermarks helps to date manuscripts to certain time periods with quite high precision and sometimes also to locate them.
Hence, by having a system that can identify watermarks effectively one can temporally and spatially localize manuscript origins.

A fundamental problem of watermark reproductions in databases is the diversity of acquisition methods (e.g. hand tracing, rubbing, and radiography) which leads to depictions of the exact same watermark in radically different ways (see Fig.~\ref{fig:queried_images}).
While being of utmost importance in watermarks, it is a common problem in handwritten historical documents in general, where also other elements could be depicted in several different ways, i.e., letters, motifs, and decorations. 
This problem is known as the \textbf{cross-depiction problem}, which strives towards identifying the same object, even when it is depicted in a variety of manners.

While there has been a significant improvement of deep learning methods for object recognition in general~\cite{krizhevsky2012,he2015,hu2017}, they have not been successfully applied to watermarks.
This is due to the issue that neural networks often fail at recognizing the abstract concept and can therefore easily be fooled by noise~\cite{szegedy2013} or abstract depictions~\cite{karpathy2014blog}.


The \textbf{main contribution} of this paper is a study of the robustness of deep learning based approaches to the cross-depiction problem on historical watermarks.
We report results by formulating the problem as two different tasks: classification (96\,\%) and similarity ranking (FPR95: 0.11). 
Our method performs better than the state-of-the-art.

\section{Related Work}
\label{toc:Related Work}

\begin{figure}[!t]
  \centering
  \subfloat[Sample from class: Fauna]%
  {\includegraphics[height=4cm]%
  {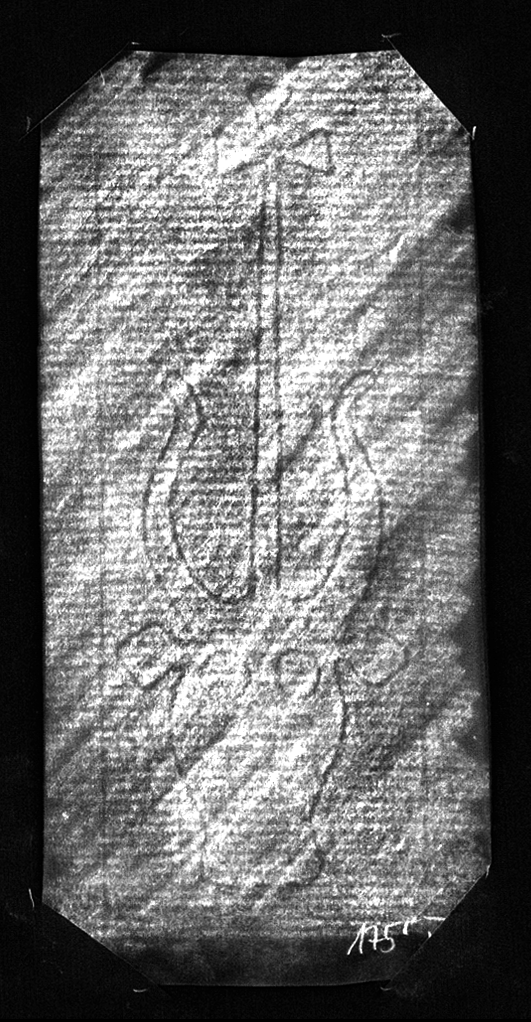}\label{subfig:watermarks_berge1}}
  \hfil
  \subfloat[Sample from correctly labeled outputs for class: Fauna]%
  {\includegraphics[height=4cm]%
  {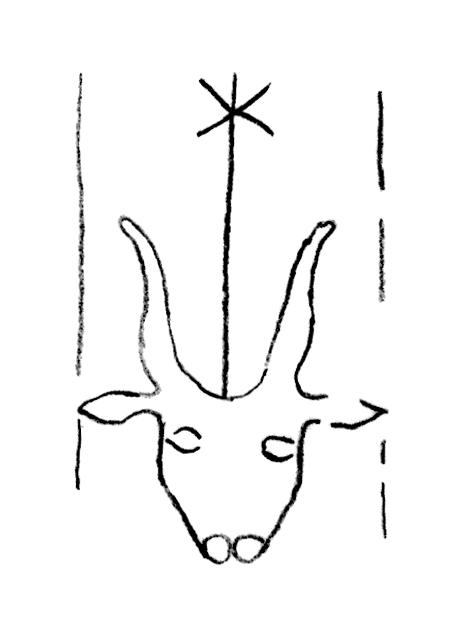}\label{subfig:watermarks_berge2}}
  
  \vfill
  \subfloat[Sample from class: Coat of arms]%
  {\includegraphics[height=4cm]%
  {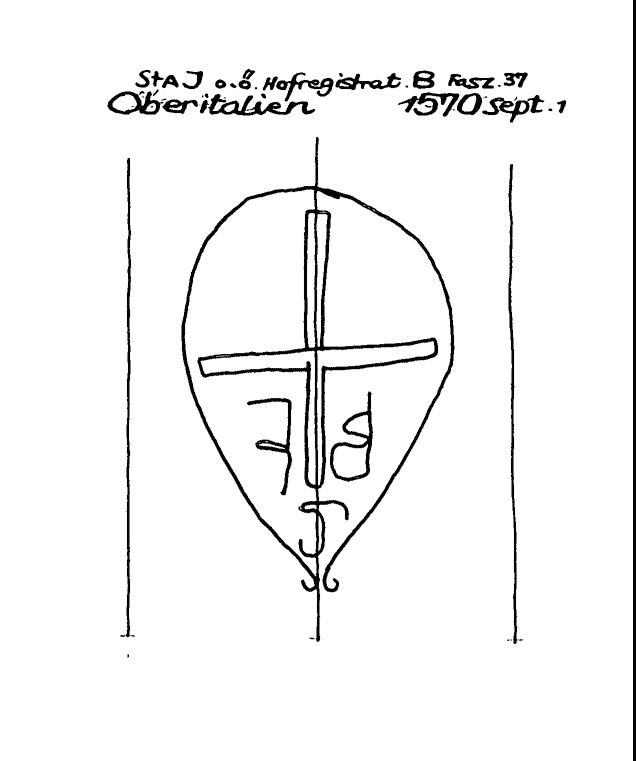}\label{subfig:watermarks_berge3}}
  \hfil
  \subfloat[Sample from wrongly labeled outputs for class: Coat of arms]%
  {\includegraphics[height=4cm]%
  {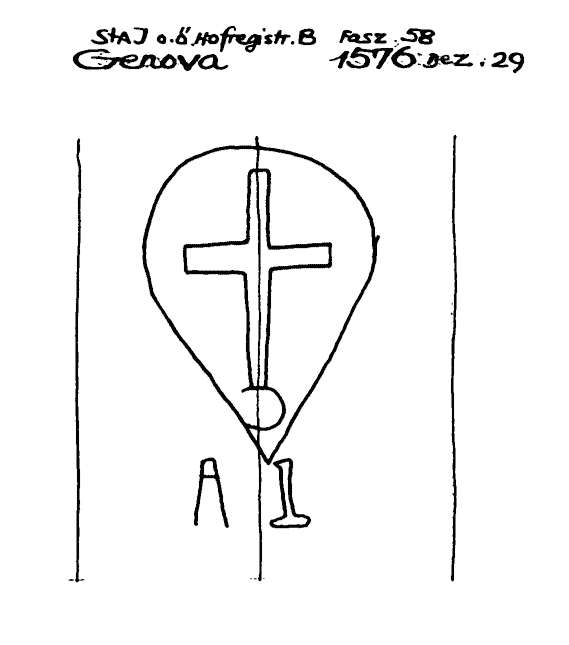}\label{subfig:watermarks_berge4}}
  
  \caption{Samples from the watermarks dataset (a and c) juxtaposed with a right (b) and wrong (d) classification results from the network. Notice how the way (a) and (b) are depicted is significantly different yet the network gives them the same label. Therefore its surprising that samples (c) and (d) --- which are much more similar in terms of appearance --- receive two different labels from the network.}
  \label{fig:watermarks}
\end{figure}

\textbf{Cross--depiction:}
A lot of work has been done in the context of object classification, object detection and image similarity but very few of them address specifically the problem of depicting an object in different ways.
The most relevant work in our context are Cai et al.~\cite{cai2015} who raised awareness on how hard is to tackle this problem and how both traditional methods and deep learning fails to solve it, and Picard et al.~\cite{picard2016} who investigated the retrieval of paper watermarks by visual similarity by encoding small regions of the watermark using a non-negative dictionary optimized on a large collection of watermarks.
Previous approaches to automatic watermark identification \cite{rauber1996, riley2002, riley2003, brunner2006, otal2008} show that watermarks are a topic that was addressed quite early by the image retrieval research, but until today, it is not solved in a satisfying way\footnote{By this we mean solved in such a way that it can be used productively in the humanities.}.
Other work on the subject include Crowley and Zisserman~\cite{crowley2014} who attempted object retrieval in paintings, Hu and Collomosse~\cite{hu2013} using HOG descriptor for sketch based image retrieval, Ginosar et al.~\cite{ginosar2015} detecting People in Cubist Art.  


\textbf{Deep Learning:} the developments of modern deep learning techniques have led to remarkable improvements in the field of computer vision.
Back in 2009, when the ImageNet~\cite{jiaDeng2009} dataset for object recognition has been released, the top ranks were dominated by traditional heuristic methods.
Only few years later, the well-known model AlexNet~\cite{krizhevsky2012} opened the road to what will be a cascade of deep learning models which would perform better and better every year.
In fact, shortly after we have witnessed the first deep neural network surpass human-level performance~\cite{he2015} and yet these results are again significantly outperformed by the latest models~\cite{hu2017}.
The effectiveness of deep learning methods has been exploited by a growing community of researchers who constantly pushed the boundaries, not only of computer vision, but of machine learning in general. 
Despite these outstanding results, we know that neural networks do not perform vision as humans do.
There are many situations in which this difference is very clear.
For example, one can look at the diversity in the inherent nature of adversarial examples for humans and computers.
While humans can be fooled by simple optical illusions~\cite{ittelson1951}, they would never be fooled by synthetic adversarial images, which are extremely efficient into deceiving a neural network~\cite{szegedy2013, yosinski2014, evtimov2017}.
Another scenario is to look into what types of error are humans or networks more susceptible to when performing object recognition.
Deep learning models tend to make mistakes on abstract representation whereas humans are very robust towards this type of error~\cite{karpathy2014blog}.
Successfully identifying an abstract representation --- such as drawings, paintings, plush toys, statues, signs or silhouettes --- is a very simple and very common task for humans, yet machines still struggle to cope with it. 

\textbf{Image Similarity:} matching image patches via local descriptors is an important research area in computer vision due to the wide range of its application, i.e., object recognition and image retrieval. 
There is a vast literature on the subject and here we briefly describe the work most relevant to the architecture we used in our experiments.
Specifically, we adopt the triplet network model~\cite{hoffer2015, balntas2016} which has been shown to outperform two-channel networks~\cite{zagoruyko2015} and advanced application of the Siamese approach such as MatchNet~\cite{han2015} as well.

\section{Dataset}
\label{toc:Dataset}

We used a dataset provided by the watermark database Wasserzeichen Informationssystem (WZIS)\footnote{\url{https://www.wasserzeichen-online.de/wzis/struktur.php}} which contains in total $106'502$ watermark reproductions stored as RGB images of size approximately\footnote{Not all images have the same size. The numbers reported are the average over the whole dataset.} $1500\times 720$.
Most of them (around $92'000$) are hand tracings by Gerhard Piccard, who started gathering and publishing a huge watermark collection from the 1960’s~\cite{piccard1978wasserzeichenkartei}.
Although, in more recent watermark research, new reproduction techniques have also become important, such as rubbing, photography, radiography, and thermography.
The different image characteristics between tracings (pen strokes, black and white) and the other reproduction methods (less distinct shapes, grayscale) makes the task of watermark classification and recognition more difficult (e.g. notice how in Fig.~\ref{subfig:query} and ~\ref{subfig:e1} the same object is represented in two radically different ways). 
Therefore, we also included rubbings and radiography reproductions in our data set.

In the watermark research, there exist very complex classification systems for the motifs depicted by the watermarks. 
For example, the classification system used in WZIS contains 12 super-classes with around 5 to 20 sub-classes each~\cite{frauenknecht2015}.

We created three expert annotated sets containing queries with nine motif classes: bull's head, letter P, crown, unicorn, grape, triple mount, horn, tower, circle. 
The choice of these classes is either motivated by their frequency (bull's head, letter P), or by their complexity (grape, triple mount). 
The first and second test sets contain the five motif classes bull's head, letter P, crown, unicorn, and grape.
The reproduction techniques in these test sets are mixed (hand tracing, rubbing, radiography). 
The third test set contains the five motif classes bull's head, triple mount, horn, tower, and circle. 
In this test set, there are only hand tracings.

\section{Classification Task}
\label{toc:Classification Task}

\begin{figure}[!t]
	\centering
	\includegraphics[width=\columnwidth]{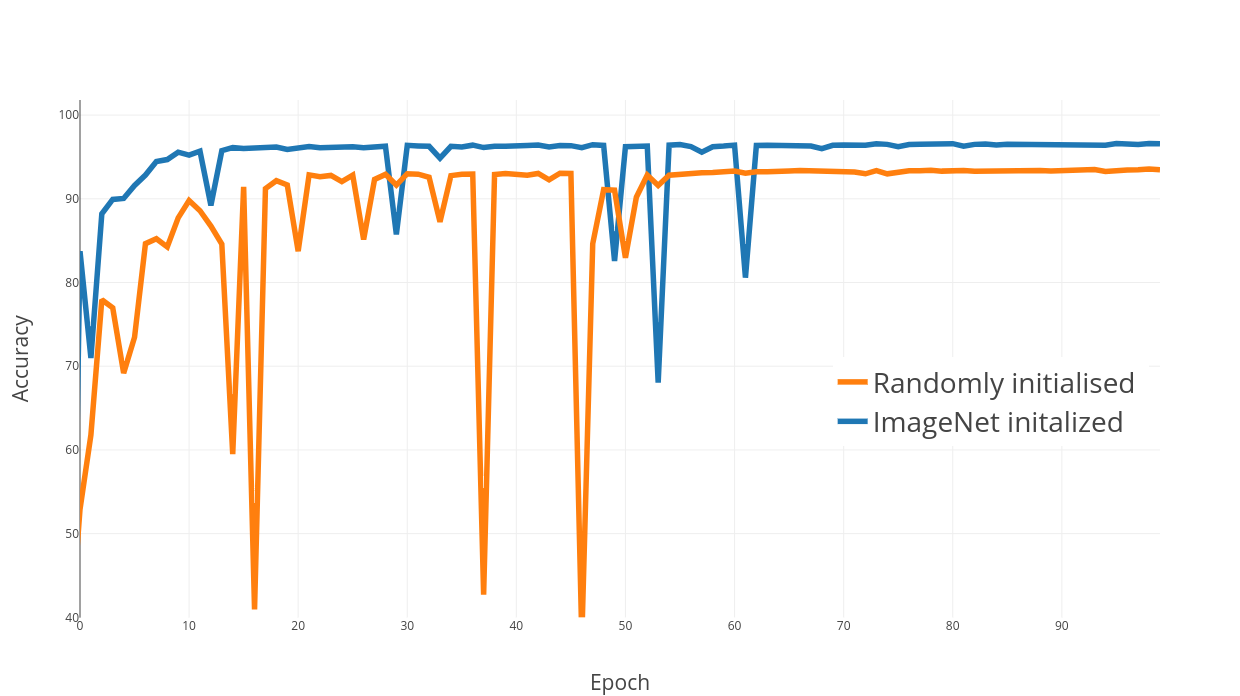}
	\caption{Comparing the effects of pre-training on classification performance on validation set.
	Orange: network initialized with random weights. 
	Blue: network initialized with ImageNet pre-trained weights.
	}
	\label{fig:compare_pretrained}
\end{figure}

The watermark classification task is an instance of the object classification task, where given an image the system has to output the correct label for the object in it.
In our context the different watermarks represent the objects to classify in the image.

\subsection{Architecture}
\label{toc:classification_architecture}

Deep neural networks are known to be difficult to train. 
In the last years different solutions have been proposed which tackle this issue by employing particular architectures to combat the gradient vanishing problem. 
Among them there are Long Short-Term Memory (LSTM)~\cite{hochreiter1997} networks and Residual Networks (ResNet)~\cite{he2016deep}.
The former is a type of recurrent neural network and uses specially tailored gate units to control the flow gradients to prevent the gradient vanishing problem.
The latter is a variant of Convolutional Neural Network (CNN) and it introduces skip connections which perform identity mapping to prevent the gradient from vanishing even in extremely deep networks.
Skip connections increase effectiveness of training on deeper network and achieves similar performance to standard networks~\cite{simonyan2014very} with less computations on shallower ones.
In this work we use an 18-layers ResNet as orignally specified on the PyTorch documentation\footnote{https://github.com/pytorch/vision/blob/master/torchvision/models/resnet.py}.

\subsection{Experimental setting}

We compare the effectiveness of using a variant of the network that has been pre-trained on the ImageNet dataset from the ImageNet Large Scale Visual Recognition Challenge~\cite{russakovsky2015} (ILSVRC) against the same model but randomly initialized. 
Afzal et al.~\cite{afzal2015deepdocclassifier} have shown that ImageNet pre-training can be beneficial for several document image analysis tasks. 

We perform all experiments using the DeepDIVA experimental framework~\cite{alberti2018deepdiva}. 
We use the Stochastic Gradient Descent optimizer to train for $100$ epochs with a standard learning rate of $0.01$. 
The images are then resized to a resolution of $224$~x~$224$ to be compatible with the expected input size of the model . 
Additionally we scale the class-wise weight updates by the inverse of the frequency of the class to prevent the network from over-fitting to the distribution of the data\footnote{This is often referred to as ``data balancing''.}. 

For the classification task we used the 12 super-classes and split the dataset in $76'947$ training images, $13'579$ for validation and $15'976$ for testing.

\subsection{Results}

The evolution of the validation accuracy during training for both the pre-trained and the randomly initialized networks can be seen in Fig.~\ref{fig:compare_pretrained}.
The pre-trained network outperforms the random counterpart both in terms of final accuracy and stability during training (notice the magnitude and frequency of the spikes).
This observation is confirmed on test set as shown in Tab.~\ref{tab:classification_perf}), with a difference of approximately $3\%$ between the two networks.

\begin{table}[]
\centering
\caption{Classification performance of ResNet18}
\label{tab:classification_perf}
\begin{tabular}{@{}cccc@{}}
\toprule
Metric: Accuracy &Training set & Test set & Validation set\\
 \midrule
Randomly Init.& 100\,\%&93.61\,\%& 93.47\,\% \\
\\ImageNet Init. & 100\,\%& 96.42\,\% & 96.58\,\%\\

\bottomrule
\end{tabular}
\end{table}

\section{Similarity Matching Task}
\label{toc:Similarity Matching Task}

\begin{figure*}[!t]
\centering
\begin{tabular}{ccccccc}
    \toprule
    Query Image & R1 &  R2 &  R3 &  R4 &  R5 &  R6 \\
    
    \midrule
    \includegraphics[height=2.99cm]{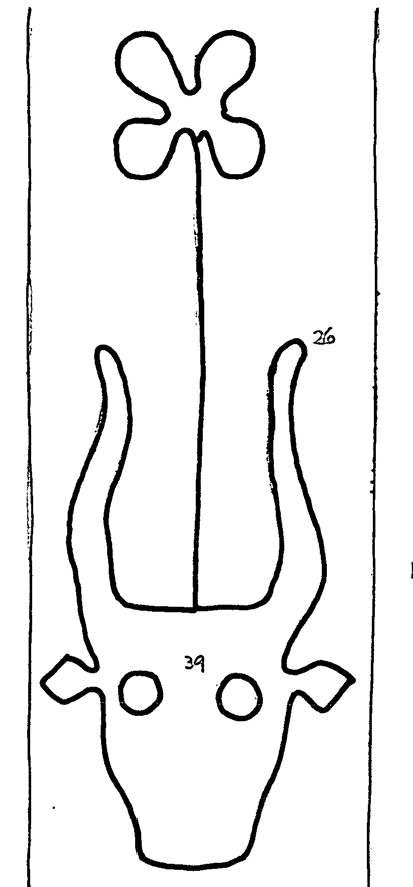} &
    \includegraphics[height=2.99cm]{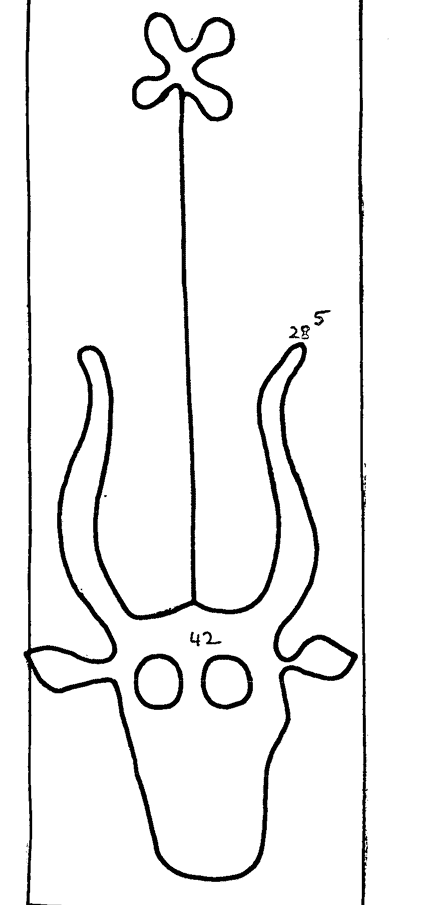} &
    \includegraphics[height=2.99cm]{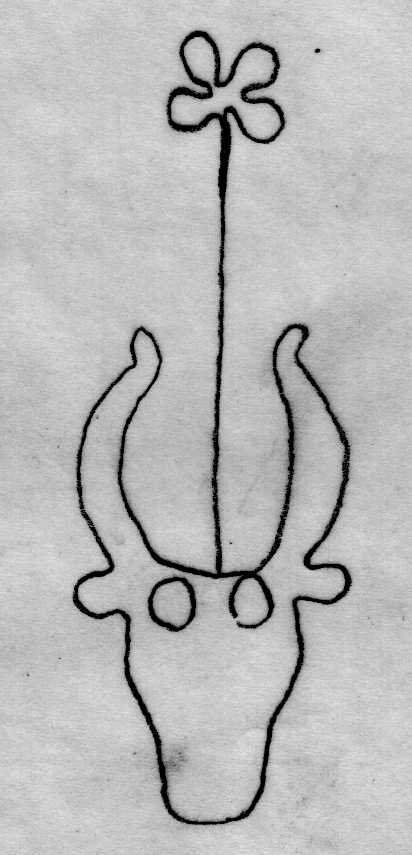} &
    \includegraphics[height=2.99cm]{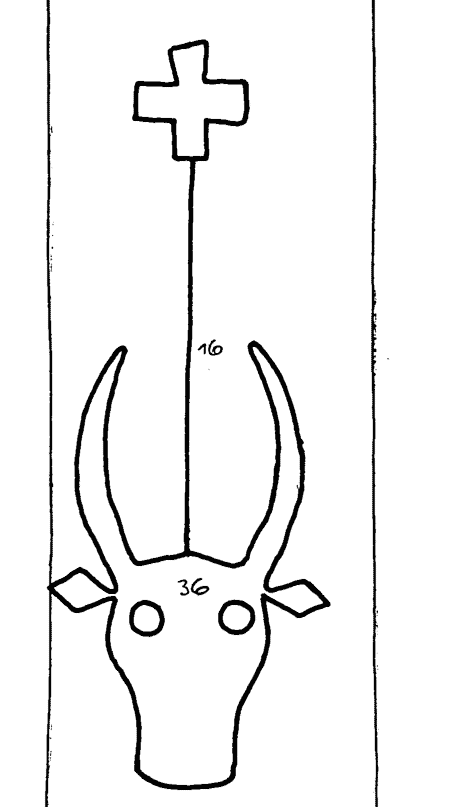} &
    \includegraphics[height=2.99cm]{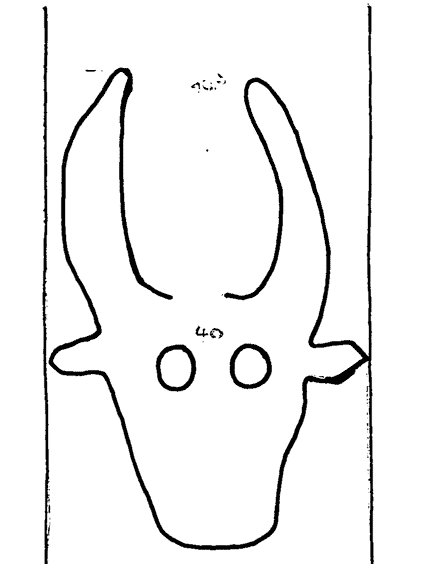} &
    \includegraphics[height=2.99cm]{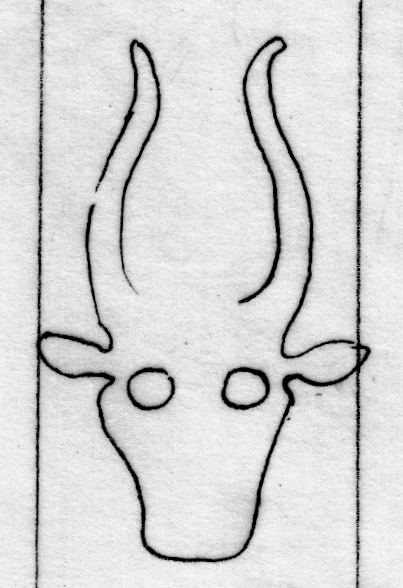} &
    \includegraphics[height=2.99cm]{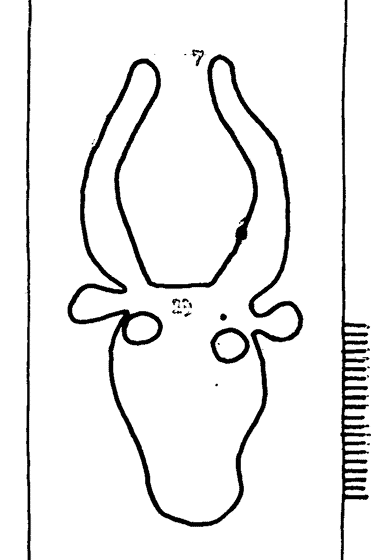} \\
    \midrule
    
    Our approach  &
    \raisebox{-.5\height}{\includegraphics[height=2.99cm]%
    {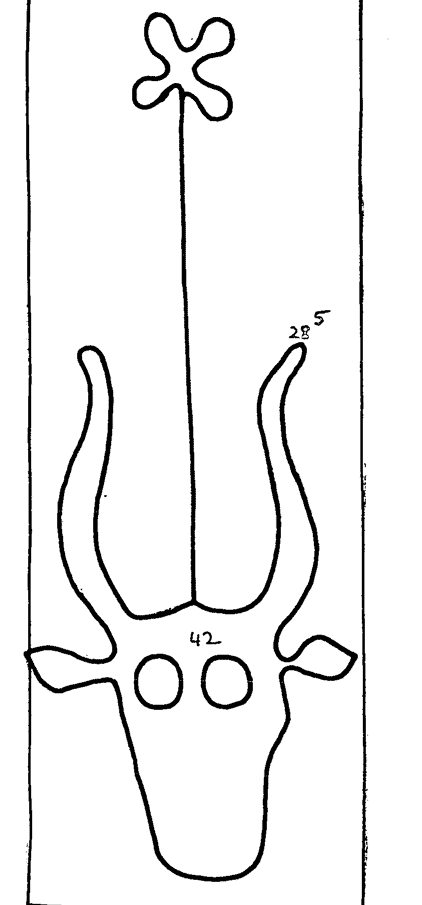}} & 
    \raisebox{-.5\height}{\includegraphics[height=2.99cm]%
    {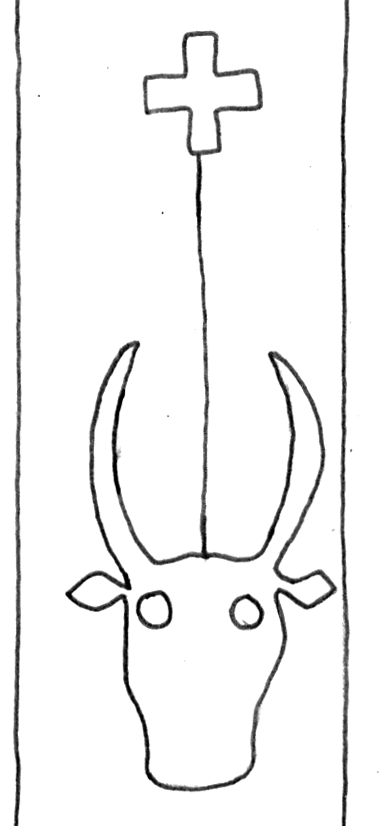}} &
    \raisebox{-.5\height}{\includegraphics[height=2.99cm]%
    {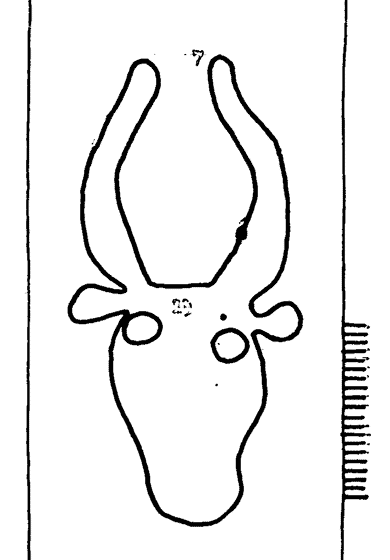}} &
    \raisebox{-.5\height}{\includegraphics[height=2.99cm]%
    {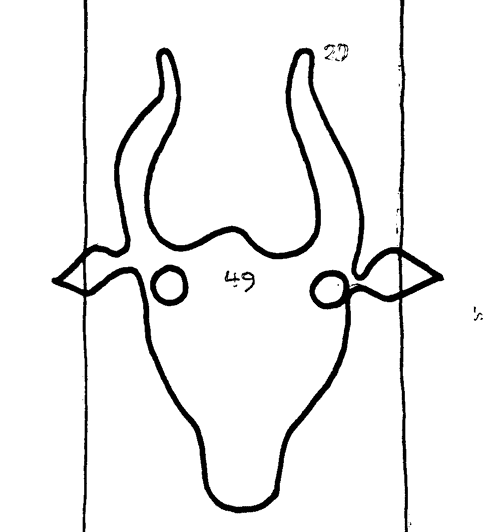}} &
    \raisebox{-.5\height}{\includegraphics[height=2.99cm]%
    {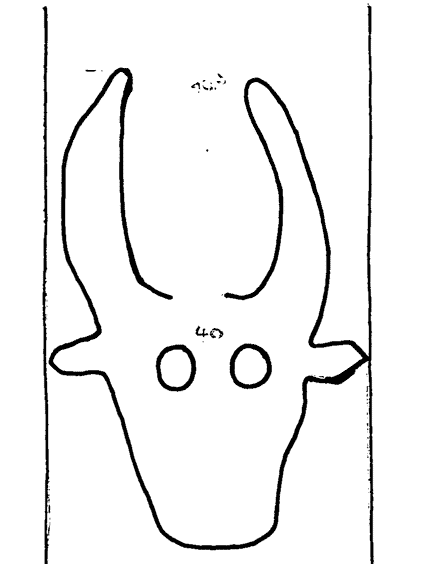}} &
    \raisebox{-.5\height}{\includegraphics[height=2.99cm]%
    {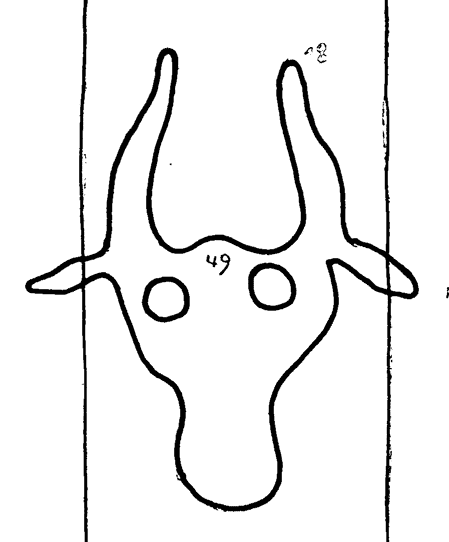}} \\ \\
    
    IOSB &
    \raisebox{-.5\height}{\includegraphics[height=2.99cm]%
    {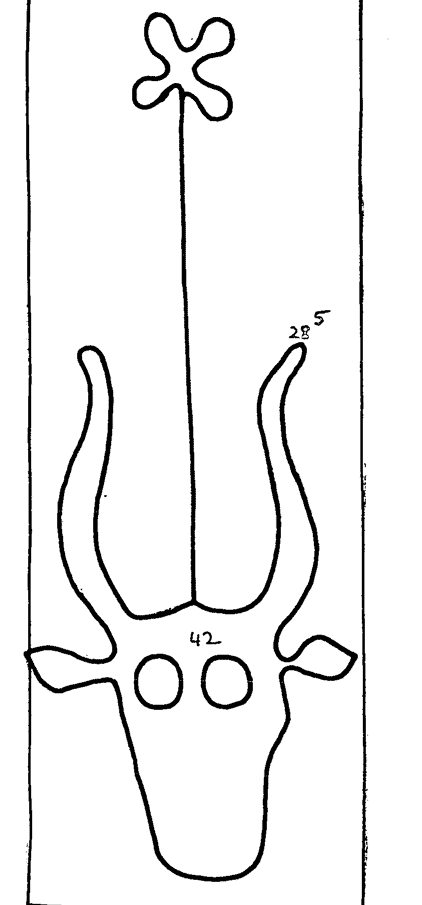}} &
    \raisebox{-.5\height}{\includegraphics[height=2.99cm]%
    {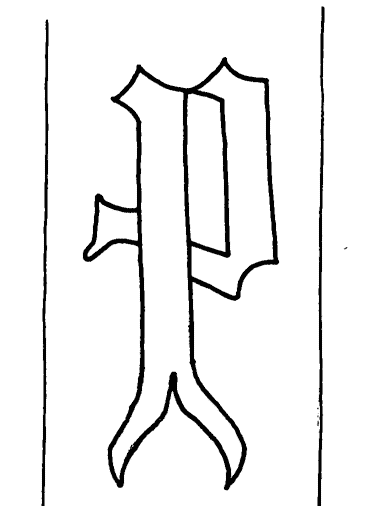}} &
    \raisebox{-.5\height}{\includegraphics[height=2.99cm]%
    {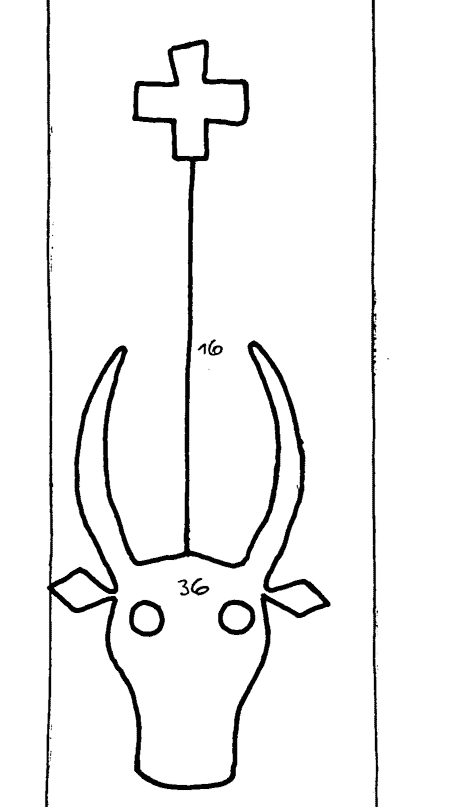}} &
    \raisebox{-.5\height}{\includegraphics[height=2.99cm]%
    {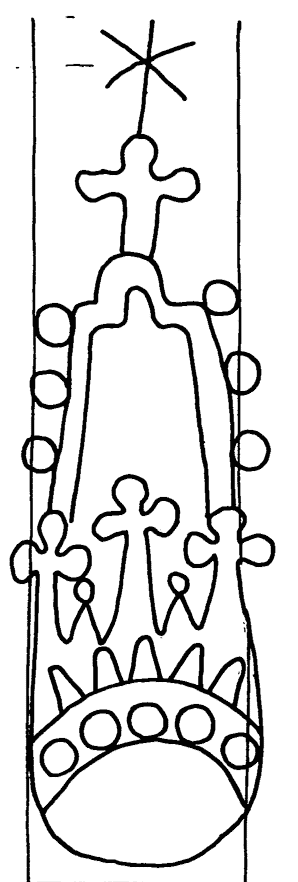}} &
    \raisebox{-.5\height}{\includegraphics[height=2.99cm]%
    {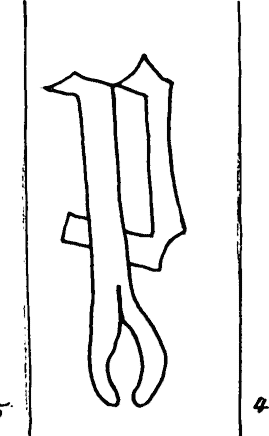}} &
    \raisebox{-.5\height}{\includegraphics[height=2.99cm]%
    {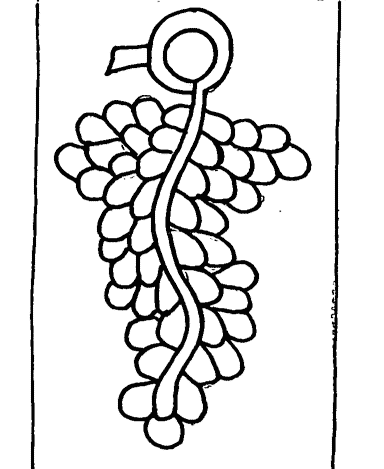}} \\ \\
    
    LIRe &
    \raisebox{-.5\height}{\includegraphics[height=2.99cm]%
    {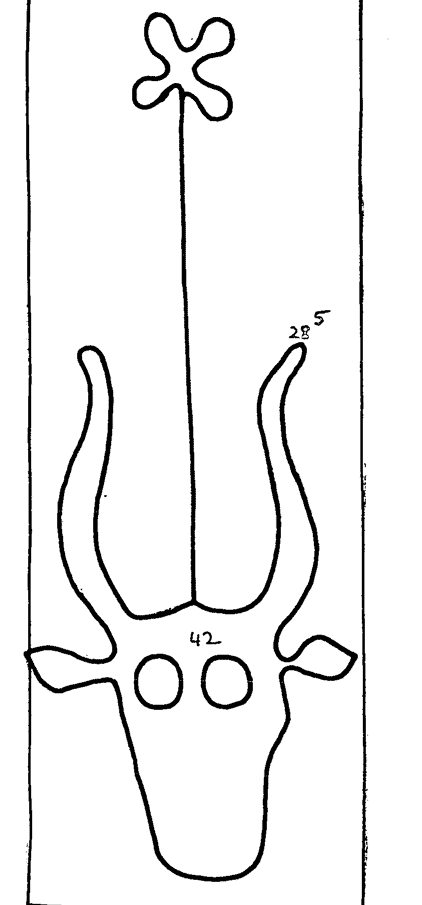}} &
    \raisebox{-.5\height}{\includegraphics[height=2.99cm]%
    {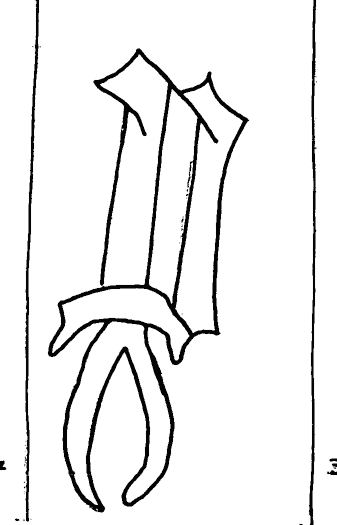}} &
    \raisebox{-.5\height}{\includegraphics[height=2.99cm]%
    {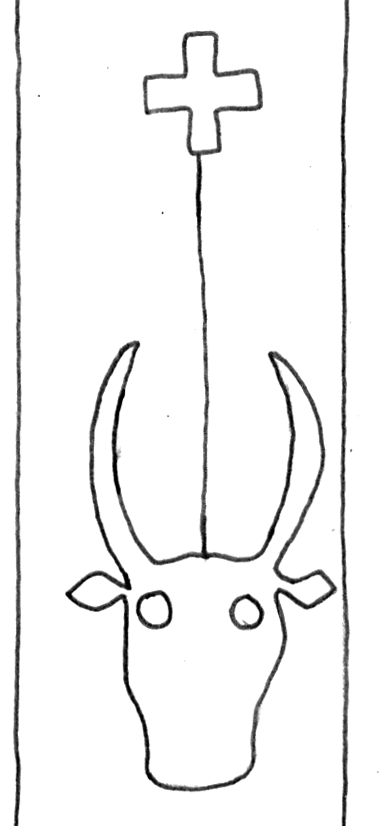}} &
    \raisebox{-.5\height}{\includegraphics[height=2.99cm]%
    {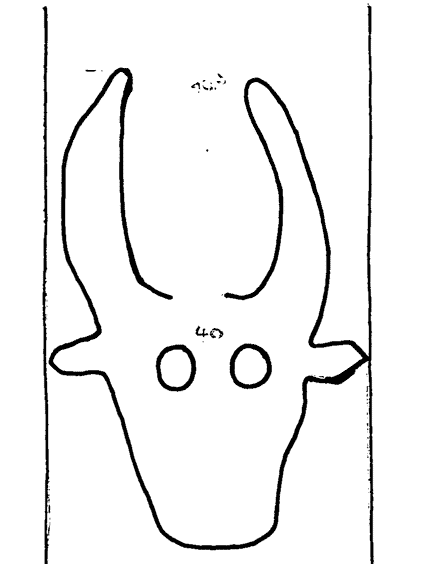}} &
    \raisebox{-.5\height}{\includegraphics[height=2.99cm]%
    {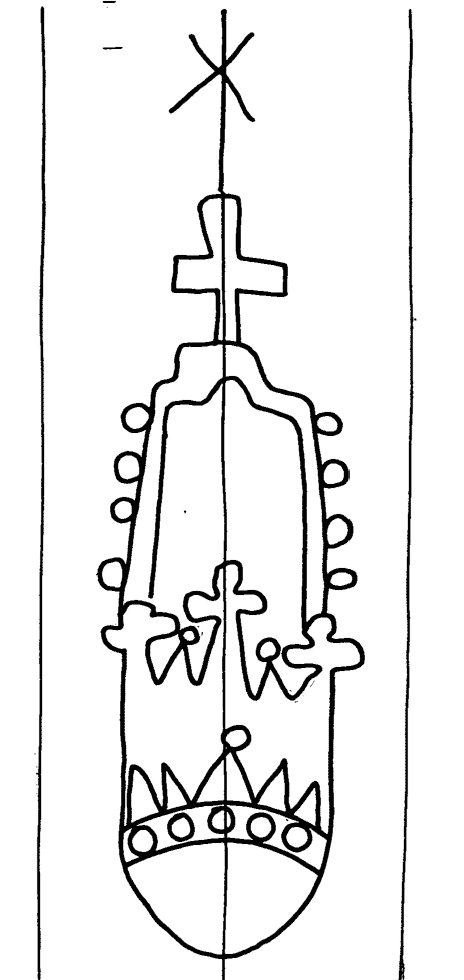}} &
    \raisebox{-.5\height}{\includegraphics[height=2.99cm]%
    {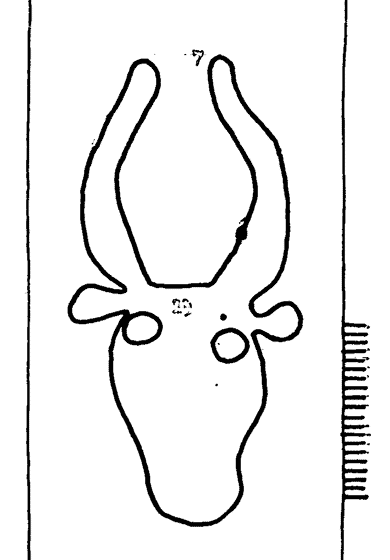}} \\
    \bottomrule
\end{tabular}

\caption{The top row shows a query image belonging to class Fauna-Bull's head and the first 6 results annotated by experts (used as ground truth for evaluation). The second row shows the results retrieved by our approach when pre-trained for classification. Notice how despite the images are not the very same, they all belong to the correct class. The third row shows the results retrieved by the IOSB system \cite{manger2012} and the last row those of the LIRe system \cite{lux2011}. In these cases, occasionally some of the retrieved images are not only dissimilar to the query image but also belonging to another class e.g. second and fifth columns.}
\label{fig:competitors}
\end{figure*}

\begin{figure}[!t]
  \vfill
  \subfloat[Randomly initialized network]%
  {\includegraphics[width=.48\columnwidth,height=4cm]%
  {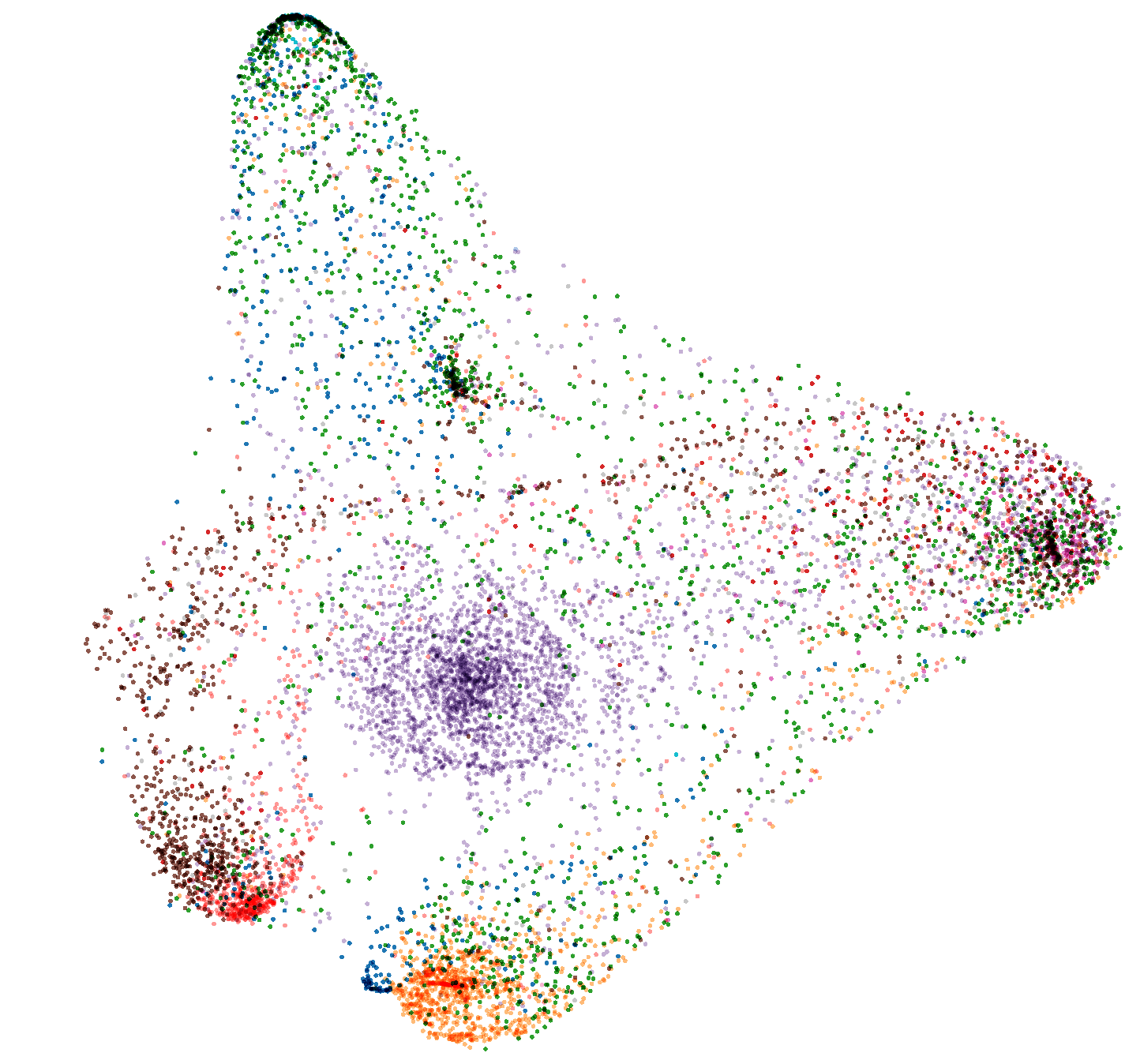}\label{subfig:e0}}
  \hfil
  \subfloat[Classification pretrained network]%
  {\includegraphics[width=.48\columnwidth,height=4cm]%
  {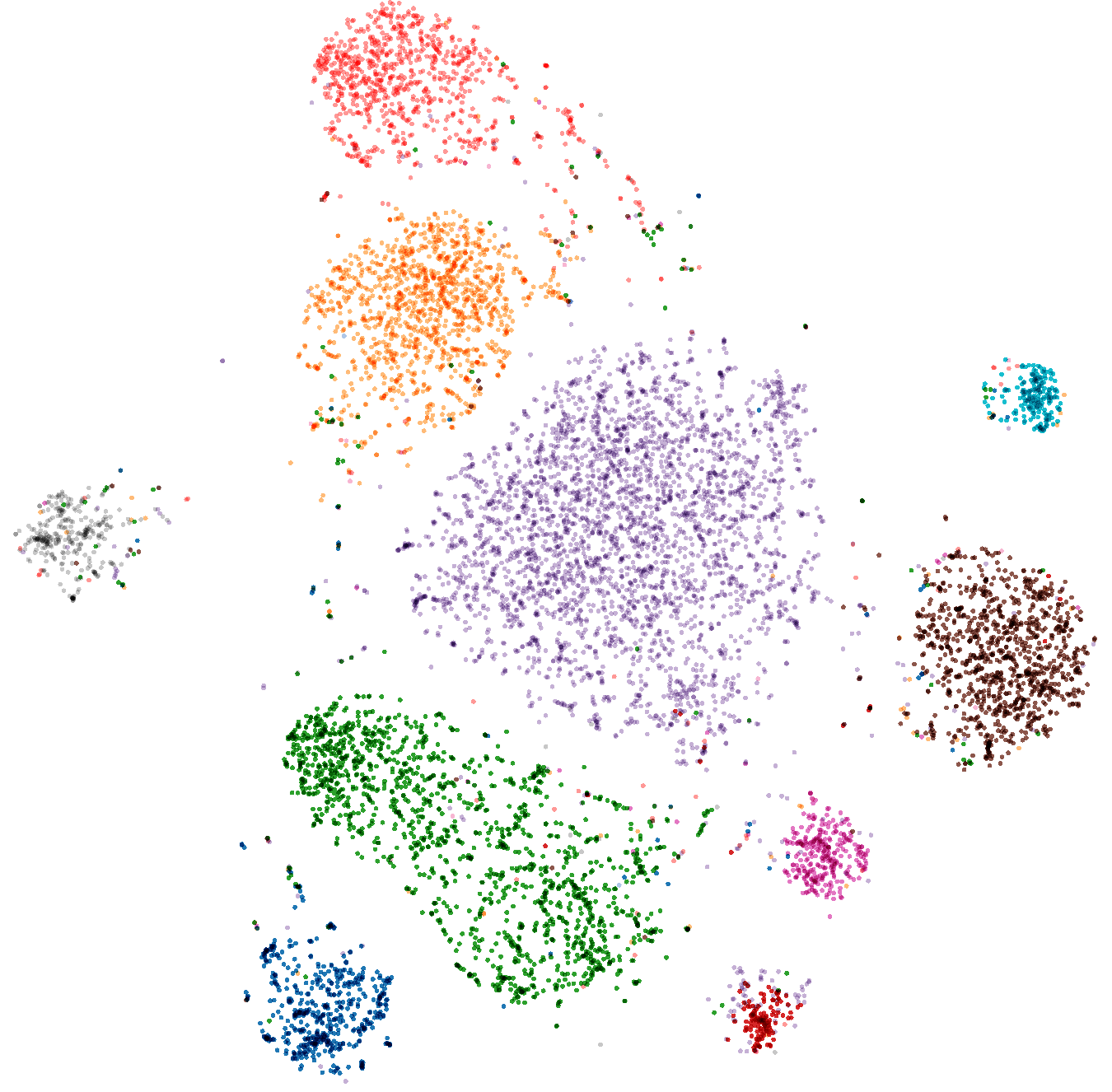}\label{subfig:e10}}
  \caption{T-Distributed Stochastic Neighbor Embedding (T-SNE)~\cite{maaten2008visualizing} visualization of the validation set images in the latent embedding space. Different colors denotes samples belonging to different classes.}
  \label{fig:embeddings}
\end{figure}

\begin{figure}[!t]
	\centering
	\includegraphics[width=\columnwidth]{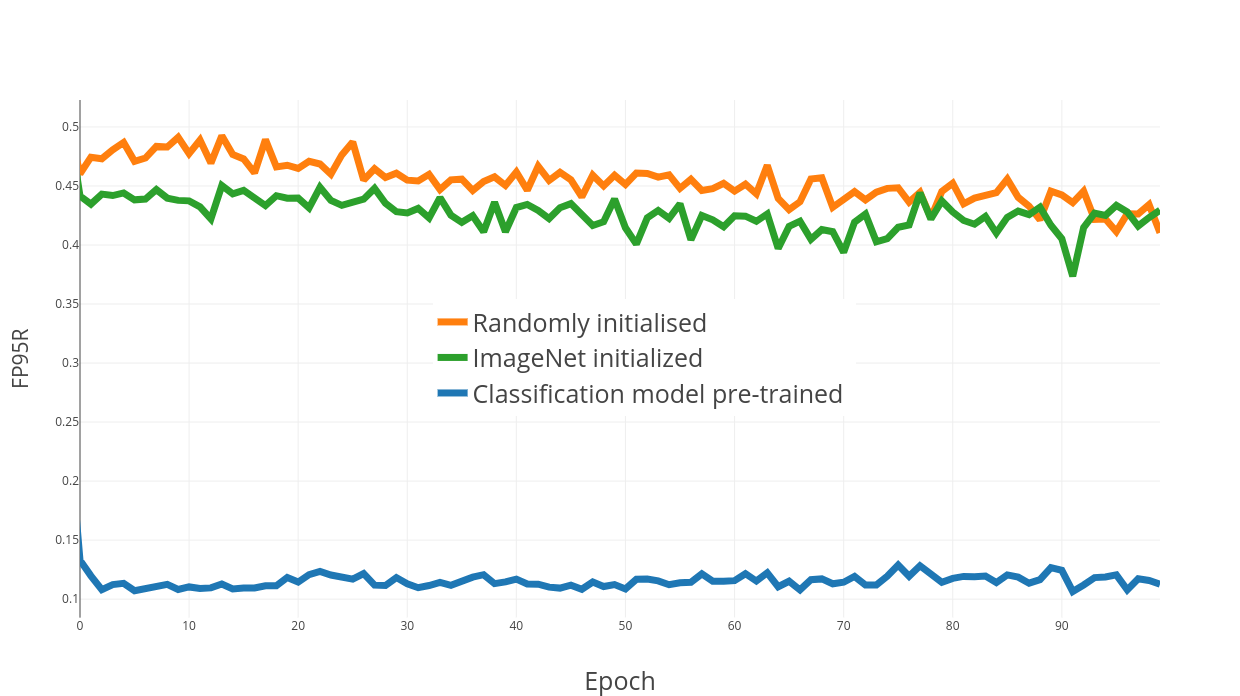}
	\caption{False Positive at 95\% Recall scores on the validation set for similarity matching task.}
	\label{fig:effect_of_pretraininig_for_classification}
\end{figure}

The similarity matching task can be formulated such that for a given query, a system is required to return the most similar images to it.
In other cases the ground truth of how similar two images are can be inferred at the creation of the dataset, e.g. given a query image, other images of the same subject are considered to be inherently similar whereas images of other subjects are considered to be dissimilar.
In our situation this not possible because there are watermarks which look very similar despite belonging to two different classes and the converse is also true: there are watermarks which look very different yet they are from the same class.
This effect is particularly strong if we were to consider a finer grain of classes rather than the 12 used for classification (see details in Section~\ref{toc:Dataset}).
Additionally, it is difficult to judge the similarity of multiple images belonging to the same class with a metric such that they can be sorted.
For this reason, to evaluate the quality of our similarity matching we use queries for which the desired output has been defined by  experts in the field of humanities. 
One of such query can be seen in Fig.~\ref{fig:competitors}.
Generating these queries is very time consuming and the expertise required to do so prohibits using tools such as crowd-sourcing as  similarly done in computer vision.

\subsection{Architecture}

For this task we use the very same network we previously used for the classification task with a minor modification on the last layer.
The network is altered such that it no longer outputs class labels but rather an embedding of the input image in a space where similar images will have close\footnote{Distance measure with Euclidean distance.} embeddings. 
We therefore ablate the last layer of the network and replace it with a fully connected layer of size 128.

\subsection{Experimental setting}

Similarly to what done in classification we study the effect of different initialization for the network.
This time instead of comparing ImageNet pre-trained networks against a randomly initialized one we compare it additionally against our best model obtained after training for classification.
The hypothesis is that a network which performs so well for classifying the watermarks might have learned some specific filters for this dataset and therefore can be either trained faster or perform better than one pre-trained only on ImageNet\footnote{The input domain of ImageNet is significantly different than the one of our watermark dataset.}.

We train the network to minimize the margin ranking loss as proposed in~\cite{wang2014}.        
           
\subsection{Results}

We report the results in terms of false positive rate at 95\% true positive rate (FPR95).
The FPR95 is a well established metric in the context of similarity matching and should be interpreted as the lower the better, with optimal score 0.

The results for the randomly initialized, ImageNet pre-trained, and classification pre-trained networks are reported in Tab.~\ref{tab:similarity_performances}.
The classification pre-trained network outperforms the other networks by a significant margin.
This suggests that the additional data seen during the classification training is helpful for later use when being trained for similarity matching.
This can also be seen in Fig.\ref{fig:embeddings}, where the T-SNE visualization of the embeddings produced by the randomly initialized model display less clustering tendencies than that of the embeddings produced by the classification pre-trained model.

\begin{table}[]
\centering
\caption{Similarity matching performance }
\label{tab:similarity_performances}
\begin{tabular}{@{}ccc@{}}
\toprule
 & FPR95\\\midrule
 Randomly initialized ResNet18 & 0.41
 \\ImageNet pretrained ResNet18 & 0.42
 \\Classification pretrained ResNet18 & \textbf{0.11} \\ 
\bottomrule
\end{tabular}
\end{table}

\section{Discussion and Analysis}
\label{toc:Discussion}

Considering the different nature of the two tasks, classification and similarity matching, it is difficult to make an objective statement about whether we have been more successful in one or the other. 
This is due to two main issues.
First, there is not much research yet for deep learning applied in similar areas.
Second, and more important, the acquisition of large datasets labelled by experts is very time consuming (see Section~\ref{toc:Similarity Matching Task}).
As there are no publicly available benchmark datasets yet, it makes it difficult to compare to previous work.

In order to approach qualitative evaluation of  our approach, we compare it with other existing systems, IOSB~\cite{manger2012} and LIRe~\cite{lux2011}.\footnote{Since the other tools have some technical limitations, we cannot provide a quantitative comparison as of today.}
We evaluated them on the same expert-ranked test queries.
Fig.~\ref{fig:competitors} shows a sample query and the expert annotation solution in the top row.
Below, the results of the different approaches are shown, ranked by the similarity reported by the systems.
Notice how our approach performs visibly better than the others.
Fig.~\ref{fig:queried_images} suggests that our triplet-network based approach is able to solve the cross-depiction problem rather nicely.
The target images appear in the top three results, just the ranking is not perfect. 
Fig.~\ref{fig:competitors} supports the statement that our approach finds more similar images than existing tools. 
However, a closer look reveals that the expert's results R2 and R5 do not appear among the top candidates.
This might be due to two issues: (i) the images are with a non uniform background, (ii) the images are free hand-drawn sketches while the others are traced.
We plan to investigate the reasons further and can apply: (i) binarization or filtering techniques, (ii) data augmentation, where the lines are slightly deformed.

\section{Conclusion}
\label{toc:conclu}

We have shown that with very deep models can be robust enough even in the context of the cross-depiction problem.
We measured their performance on two different tasks: classification and similarity rankings using a dataset provided by the WZIS watermark database.
The results are promising as we achieve a classification accuracy on the test set of 96\,\%, and a similarity performance of 0.11 FPR95.
These results outperform state-of-the-art methods by a significant margin.
Future work should investigate the generality of our findings in other datasets and with more fine-grained classes. 

\section*{Acknowledgment}
The work presented in this paper has been partially supported by the HisDoc III project funded by the Swiss National Science Foundation with the grant number $205120$\textunderscore$169618$.

\bibliographystyle{IEEEtran}
\bibliography{biblio}

\end{document}